\documentclass[twocol]{ametsocV6.1}

\title{Machine Learning Estimation of Maximum Vertical Velocity from Radar}

\authors{Randy J. Chase\correspondingauthor{Randy J. Chase, randy.chase@colostate.edu}\aff{a,b,c}, Amy McGovern\aff{a,b,c}, Cameron R. Homeyer\aff{b}, Peter J. Marinescu\aff{d}, Corey K. Potvin\aff{b,c,e}}

\affiliation{\aff{a}{School of Computer Science, University of Oklahoma, Norman OK USA}\\
\aff{b}{School of Meteorology, University of Oklahoma, Norman OK USA}\\
\aff{c}{NSF AI Institute for Research on Trustworthy AI in Weather, Climate, and Coastal Oceanography, University of Oklahoma, Norman OK USA}\\
\aff{d}{Department of Atmospheric Science, Colorado State University, Fort Collins CO USA}\\
\aff{e}{National Severe Storms Laboratory, Norman, Oklahoma}\\
}

\abstract{The quantification of storm updrafts remains unavailable for operational forecasting despite their inherent importance to convection and its associated severe weather hazards. Updraft proxies, like overshooting top area from satellite images, have been linked to severe weather hazards but only relate to a limited portion of the total storm updraft. This study investigates if a machine learning model, namely U-Nets, can skillfully retrieve maximum vertical velocity and its areal extent from 3-dimensional gridded radar reflectivity alone. The machine learning model is trained using simulated radar reflectivity and vertical velocity from the National Severe Storm Laboratory's convection permitting Warn on Forecast System (WoFS). A parametric regression technique using the sinh-arcsinh-normal distribution is adapted to run with U-Nets, allowing for both deterministic and probabilistic predictions of maximum vertical velocity. The best models after hyperparameter search provided less than 50$\%$ root mean squared error, a coefficient of determination greater than 0.65 and an intersection over union (IoU) of more than 0.45 on the independent test set composed of WoFS data. Beyond the WoFS analysis, a case study was conducted using real radar data and corresponding dual-Doppler analyses of vertical velocity within a supercell. The U-Net consistently underestimates the dual-Doppler updraft speed estimates by 50$\%$. Meanwhile, the area of the 5 and 10 $\mathrm{m \ s^{-1}}$ updraft cores show an IoU of 0.25. While the above statistics are not exceptional, the machine learning model enables quick distillation of 3D radar data that is related to the maximum vertical velocity which could be useful in assessing a storm's severe potential.}

\begin{document}

\maketitle

%
%
%
\statement

All convective storm hazards (tornadoes, hail, heavy rain, straight line winds) can be related to a storm's updraft. Yet, there is no direct measurement of updraft speed or area available for forecasters to make their warning decisions off of. This paper addresses the lack of observational data by providing a machine learning solution that skillfully estimates the maximum updraft speed within storms from only the radar reflectivity 3D structure. After further vetting the machine learning solutions on additional real world examples, the estimated storm updrafts will hopefully provide forecasters with an added tool to help diagnose a storms hazard potential more accurately.  

\clearpage

%
%
%

%


\section{Introduction}

Weather hazards in the United States cost billions of dollars annually \citep{NCEI2023}. A majority of the billion-dollar disaster events involve hazards directly created by convective storms (e.g., hail, tornadoes, floods, straight-line winds). Convective weather hazards are ultimately connected to the fast current of upward moving air, also known as an updraft. Despite the updraft’s importance in convective storms, an updraft’s intensity or area is not something that is reliably quantified in real time to be used as a forecasting tool for assessing storm severity.

One method to measure a storm's updraft speed is to use multiple Doppler velocity radar measurements of the same storm \citep[see chapter 12.4.3 in ][]{rauber2018radar}. Using the continuity equation and the horizontal radial velocity from the multiple radars, a 3-dimensional (3D) wind vector can be derived from which the vertical component is the vertical velocity. While multi-Doppler wind retrievals have been used for decades \citep[e.g.,][]{Doviak1976}, the operational ground based radar system in the United States \citep[i.e., NEXRAD;][]{Crum1993} is not configured in a way that is conducive for high quality multi-Doppler measurements of storm updrafts (e.g., large baselines, missing low level coverage). Furthermore, the sensitivity of multi-Doppler analyses to artifacts usually requires some manual quality control by a domain expert, making a timely real time product prohibitive \citep[see section 3 in][for an example of multi-Doppler quality control efforts]{Addison2017}. Given the lack of direct measurements of storm vertical velocity, several proxy methods have been used in a research capacity to diagnose relationships between other storm updraft characteristics and their severe hazards: overshooting tops \citep[e.g.,][]{Marion2019}, lightning \citep[e.g.,][]{Carey2019}, differential reflectivity (Zdr) columns \citep[e.g.,][]{French2021}, bounded weak echo regions (BWER) \citep[e.g.,][]{Lakshmanan2000} and mesocyclone width \citep[e.g.,][]{Sessa2020}.

Overshooting tops are convective updrafts that exceed their level of neutral thermal buoyancy (i.e., equilibrium level), creating an area of colder cloud top brightness temperatures that can be detected from imaging satellites \citep[e.g., GOES;][]{Bedka2010}. Storms that have overshooting tops have been linked to severe weather on the ground \citep[e.g.,][]{Reynolds1980,Negri1981}. More quantitatively, the width of the overshooting top (i.e., overshooting top area; OTA) has been associated with tornado properties like intensity \citep{Trapp2017} and damage rating \citep[i.e.,][]{Marion2019}. Both \citet{Trapp2017} and \citet{Marion2019} found evidence that a wider OTA, and thus a wider overall updraft, is associated with more damaging tornadoes. The key hypothesis from \citet{Trapp2017} and \citet{Marion2019}, is that a wider updraft can support a wider tornadic circulation and thus more damaging tornado. Note that the OTA method can only obtain direct information of the storm's top (Fig. 1a, yellow oval). Thus, the OTA is less informative of the updraft as a whole, including the low-level updraft (i.e., 1-km above ground level) which has been postulated as more important for tornado formation \citep{Peters2023}. 

\begin{figure*}[t]
 \centering
 \noindent\includegraphics[width=6.33in]{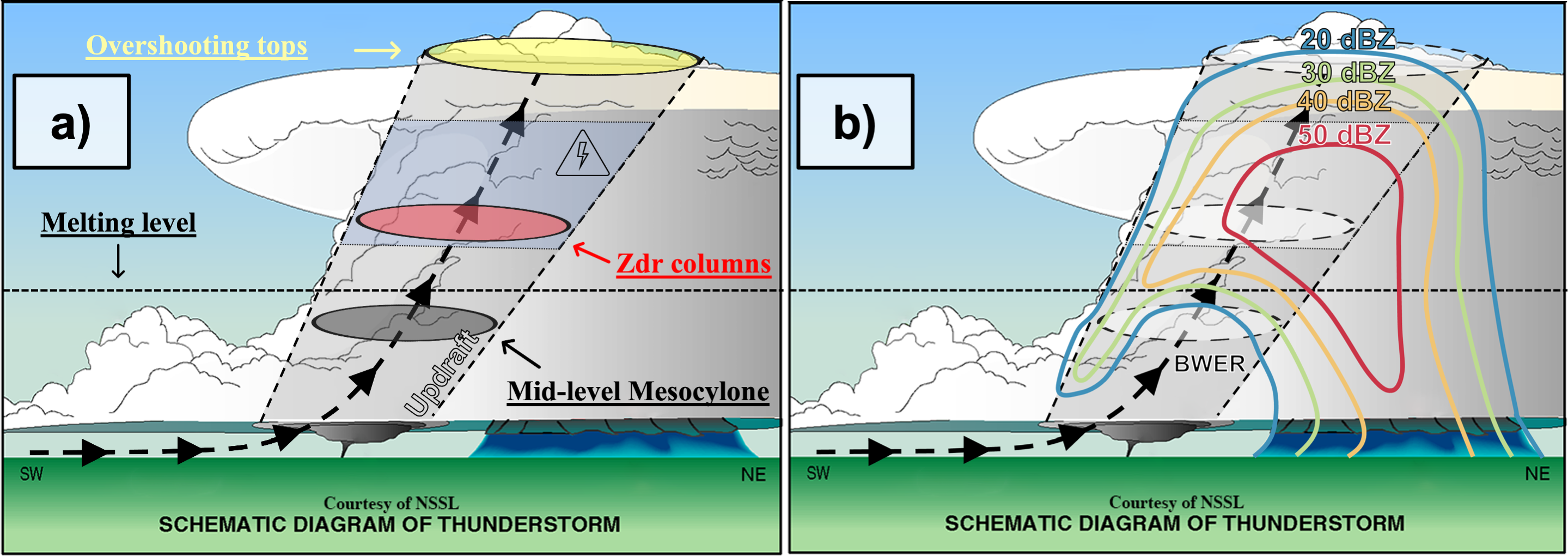}\\
 \caption{(a) Schematic summarizing past work on relating storm updraft proxies and their physical location. The black circle represents the approximate location of midlevel mesocyclones \citep[e.g.,][]{Sessa2020}, the red circle  illustrates the typical height of Zdr columns \citep[e.g.,][]{Wilson22}, the blue shaded region is the approximate region for lightning charging \citep[e.g.,][]{Deierling2008}, and the yellow circle shows the location of a overshooting top \citep[e.g.,][]{Marion2019} (b) Contours of measured radar reflectivity volumes and how it could outline the full storm updraft (inspired by \citet{chisholm1972}, reproduced in \citet{trapp2013}). The location of a Bounded Weak Echo Region \citep[BWER;][]{Browning1967} is annotated. The background supercell illustration for both (a) and (b) was provided by the National Severe Storms Laboratory.}\label{schematic}
\end{figure*}

Another method to assess convective updraft characteristics includes lightning which forms from the release of the buildup of electrical charge originating from rimed ice and pristine ice collisions within a storm's updraft. \citet{Deierling2008} and \citet{Carey2019} show that a storm's updraft volume has a clear relationship with total lightning activity. Furthermore, other studies have shown that a lightning jump, a rapid increase in lightning activity, can occur prior to severe weather \citep[e.g.,][]{Williams1999} and can be linked to the overall updraft kinematics \citep[][]{Schultz2015}. While there is evidence of a strong link between updraft volume and total lightning activity, storm microphysics is highly variable and storm to storm differences make building a general quantitative relationship is challenging. Furthermore, the best correlation between the updraft volume and lightning activity only pertains to a subset of the total updraft area (Fig. \ref{schematic}a, blue shading). 

Beyond overshooting tops and lightning, an additional proxy method for diagnosing information of a storm's updraft are Zdr columns \citep[e.g.,][]{Kumjian2008}. In some convective updrafts, storms loft raindrops above the melting level where they retain their semi-oblate shape which corresponds to a positive Zdr value (i.e., 1-4 dB) that is surrounded by ice particles which have near 0 dB Zdr. Thus, the high Zdr region outlines a portion of the updraft. \citet{French2021} and \citet{Wilson22} investigated how the Zdr column area is related to tornadoes, showing evidence that a larger Zdr column area is correlated with a more damaging tornado. Furthermore, \citet{Kuster2020} showed that Zdr columns change more clearly and earlier with updraft strength than reflectivity changes. While Zdr columns seem like a promising operational storm diagnostic, the region of the updraft that is highlighted by the Zdr column is ambiguous (e.g., Fig. \ref{schematic}a, red oval) and not always present.

Something else that can be observed by radars that are associated with updrafts is the presence of a Bounded Weak Echo Region (BWER) also known as a Weak Echo Region (WER) or a radar vault \citep[e.g.,][]{Browning1965,chisholm1970}. A WER is a relative minimum of radar reflectivity in the vicinity a storm's updraft. The lack of radar echo stems from the fast-moving air containing only small scatterers (i.e., small droplets). If the WER is bounded on both sides by discernible echo, it is then called a BWER. While these storm features have been known for a long time, their quantitative connection to the overall updraft characteristics have not been fully explored. Methods have been developed to automatically detect BWER \citep[e.g.,][]{Lakshmanan2000,Pal2006,Mahalik2019}, but have not been leveraged to understand the overall frequency and depth of BWER on an annual or spatial basis. Furthermore, the WER or BWER only characterize the lower portion of the updraft where scatterers are still small (e.g., Fig. \ref{schematic}b). Thus, they only characterize some unknown proportion of the draft and are not always present. 

A different radar-based method of measuring updraft characteristics is to measure the width of the mesocyclone. With a rotating updraft, the circulation can be easily observed on Doppler radar and then metrics like the mesocyclone width (distance between the maximum inbound and outbound velocity) can be extracted. The extracted properties of the mesocyclone have proven operationally useful for distinguishing a tornadoes potential intensity \citep{Gibbs2016,Gibbs2019,Sessa2020} providing support for the hypothesis that the updrafts characteristics are related to the tornado characteristics put forward by \citet{Trapp2017}. Similar to the previously mentioned methods, it is unclear which part of the updraft is being shown by the mesocyclone (e.g., Fig. \ref{schematic}a, black oval), but the method shows some potential in qualitatively measuring updraft characteristics if the storm has a mesocyclone. 

The proxies mentioned here for approximating updraft characteristics, velocity and width, provide incomplete information about the updraft. Furthermore, the direct measurement of updrafts from multi-Doppler measurements are not possible in real time for nowcasting applications. Thus, to fill the dearth of updraft measurements, this paper investigates if a machine learning method (i.e., U-Net) could retrieve updraft intensity and width from reflectivity data alone. More specifically, a machine learning model is trained to translate maps of radar reflectivity to maps of maximum vertical velocity. The main hypothesis is that the full 3-dimensional volume of radar reflectivity data contain structures and patterns in it that can be leveraged by a machine learning to infer the updraft characteristics (Fig. \ref{schematic}b). 

The rest of this paper is structured as follows: Section 2 discusses the datasets used in training and evaluating the machine learning method. Section 3 mentions the specifics of the methods used to engineer a dataset usable for machine learning and the machine learning details. Note that Section 3 also contains some improvements to the machine learning method used here. Section 4 discusses the primary results of this paper and Section 5 summarizes and concludes the manuscript.

\section{Data}

A goal of this research is to provide a near-real time estimate of maximum vertical velocity inferred from observed radar data. Ideally, the machine learning method would be trained using high-quality physics-based retrievals of vertical velocity from radar data. While many multi-Doppler wind retrievals exist \citep[e.g.,][]{Stechman2016,Addison2017,Stechman2020}, multi-Doppler experiments rarely use the same radar frequencies (e.g., Ka-, X-, S- band), observing platform (e.g., Doppler on Wheels, NOAA P-3 aircraft) and grid-spacing (e.g., 500 m, 1 km etc.) which makes obtaining a large standardized machine learning dataset challenging. As an alternative, this paper trains a machine learning method using convective allowing numerical weather prediction model simulations where reflectivity and vertical velocity exist on the same grid (i.e., image). More specifically, the data used to train the machine learning is from the National Severe Storms Laboratory's Warn on Forecast System \citep[WoFS;][]{Stensrud2009,Jones2020}. After training on synthetic radar data, the machine learning model is evaluated on a case study from The Colorado State University Convective CLoud Outflows and UpDrafts Experiment \citep[C\textsuperscript{3}LOUD-Ex;][]{vandenHeever2021,Marinescu2020} where multi-Doppler measurements are available for comparison.

\subsection{Training Domain: NSSL Warn on Forecast System}
The Warn on Forecast System is an ensemble of the Weather and Forecasting Research (WRF) model (36 data assimilation members, 18 forecast members) with a rapid data assimilation cycle (15 mins) and convective allowing horizontal grid-spacing (3 km). The ensemble consists of variations of physics parameterizations, namely the boundary layer scheme, surface physics and the radiation schemes \citep[see Table 2 in ][]{Skinner2018}. The variety in schemes and availability of numerous simulations of various severe weather events make the WoFS dataset attractive for training a machine learning model. 

Note that NWP simulations of convective updrafts are imperfect. For example, \citet{Varble2014}, \citet{Marinescu2016} and \citet{Fan2017} compared NWP simulations of storms to radar estimates of updrafts and the weather models consistently overestimated the magnitude of updrafts. Furthermore, the choice of microphysics scheme can bias simulated reflectivity structures \citep[e.g.,][]{Fan2017,Morrison2015} since the radar reflectivity is calculated from the bulk hydrometers prediction and assumes Rayleigh scatterers \citep[see ][ for an example of how reflectivity is calculated in WRF]{Min2015}. While these are clear limitations of the training dataset used here, the hypothesis is that since the NWP is physics informed (i.e., contains trusted physical equations), the resulting machine learning model trained from NWP data should also be physics informed. While the machine learning model might not be overly accurate (within 1 $\mathrm{m \ s^{-1}}$), the expectation for this tool is to be well correlated with the true value and be used more as a storm diagnostic (e.g., will this storm produce severe weather) by forecasters.

\subsection{Transfer Domain: GridRad}
The trained machine learning model requires radar reflectivity on a regular 3-dimensional grid. For the case study evaluation we choose to use the GridRad Severe dataset \citep{Murphy2023}. GridRad Severe is a 3-dimensional gridded radar product created from NEXRAD. The horizontal grid-spacing of GridRad is 0.02 degree x 0.02 degree (1.5 by 1.5 km) while the vertical grid-spacing is 0.5 km vertical resolution up to 7 km above mean sea level (MSL) and 1 km from 7 to 22 km MSL. The 3-dimensional analysis is conducted for every 5 min of data and the analysis is run on approximately 100 severe weather days per year over the 2010 - 2019 time period. For this paper only 26 May 2017 is used. 

\subsection{Transfer Domain: C\textsuperscript{3}LOUD-Ex Data}
Out of the C\textsuperscript{3}LOUD-Ex field campaign we use 26 May 2017, where dual-Doppler measurements between the Denver WSR88D KFTG radar and the Colorado State CHILL radar facility \citep{Brunkow2000} in Greeley, Colorado are available. More specifically we use the dual-Doppler analysis from \citet{Marinescu2020} where two dual-Doppler techniques were used, namely Spline Analysis at Mesoscale
Utilizing Radar and Aircraft Instrumentation \citep[SAMURAI;][]{Bell2012} and the Custom Editing and Display of
Reduced Information in Cartesian space \citep[CEDRIC;][]{miller1998}. Both the SAMURAI and CEDRIC dual-Doppler analyses were analyzed on 1-km horizontal grid-spacing. For more detailed information on the data used see \citet{Marinescu2020}. 

\section{Methods}

\subsection{Dataset Engineering}

\subsubsection{WoFS}

\begin{figure*}[t]
 \centering
 \noindent\includegraphics[width=5.5in]{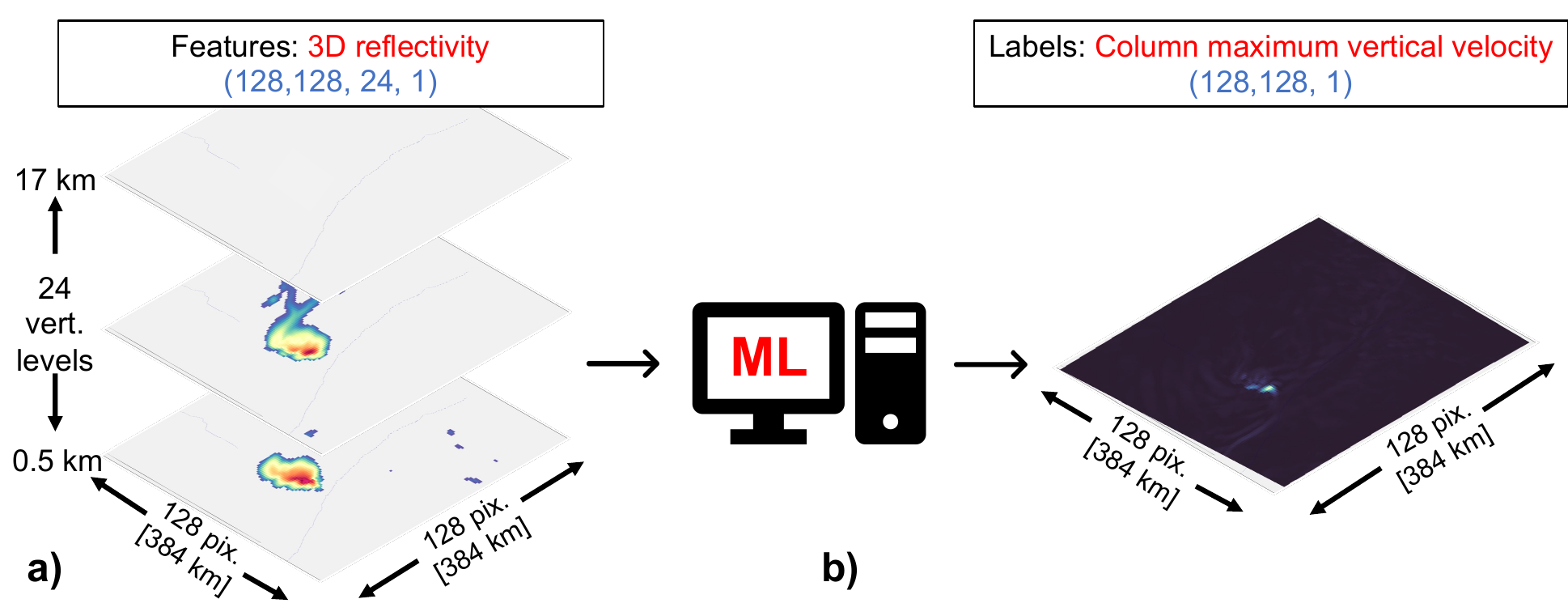}\\
 \caption{Training data schematic. (a) the 3d slices of reflectivity used for the input to the machine learning middle (center). (b) column maximum vertical velocity. Tensor shapes are found in the titles.} \label{features_labels}
\end{figure*}

Several steps are taken to prepare the original WoFS output files for machine learning. The first step is to reduce the amount of data that are available (42 TB total of data). For each day that WoFS was run, there are 18 forecast members. For each one of those members there are several initialization times (e.g., 2200, 2230, etc.) which then have forecast times out three or six hours with five minute time steps (36 files per initialization time and member). For each day, ensemble member, initialization time and forecast time, one 128 pixel by 128 pixel by 24 pixel matrix (384 km by 384 km by 17 km) of radar reflectivity data is randomly sliced out of the total domain (Fig. \ref{features_labels}a). The dimensions were chosen specifically to allow for at least three maxpooling layers within a U-Network \citep[U-Net;][]{Ronneberger2015} architecture (128 is divisible by two three times which is required for U-Net 3+). Furthermore, the 384km by 384km is a large enough domain to capture a large portion of storms, but not too large such that training on the images would be too slow (i.e., larger images require larger RAM and are slower to train/use). Other image sizes were not tested here. The corresponding maximum vertical velocity of the column (i.e., the maximum across all heights in WOFS for each grid point) is also extracted on the same 128 by 128 grid to generate the labels (Fig. \ref{features_labels}b). Maximum vertical velocity is chosen as a starting point as opposed to translating the 3-dimensional radar data to the 3-dimensional vertical velocity data, which would be a much more complex task. Only samples where there is at least one pixel of 10 $\mathrm{m \ s^{-1}}$ (or greater) are kept so maps without any convection are removed;  these maps could otherwise bias the machine learning model toward learning the trivial solution of predicting 0 $\mathrm{m \ s^{-1}}$ everywhere. 

\begin{figure}[t]
 \centering
 \noindent\includegraphics[width=2.5in]{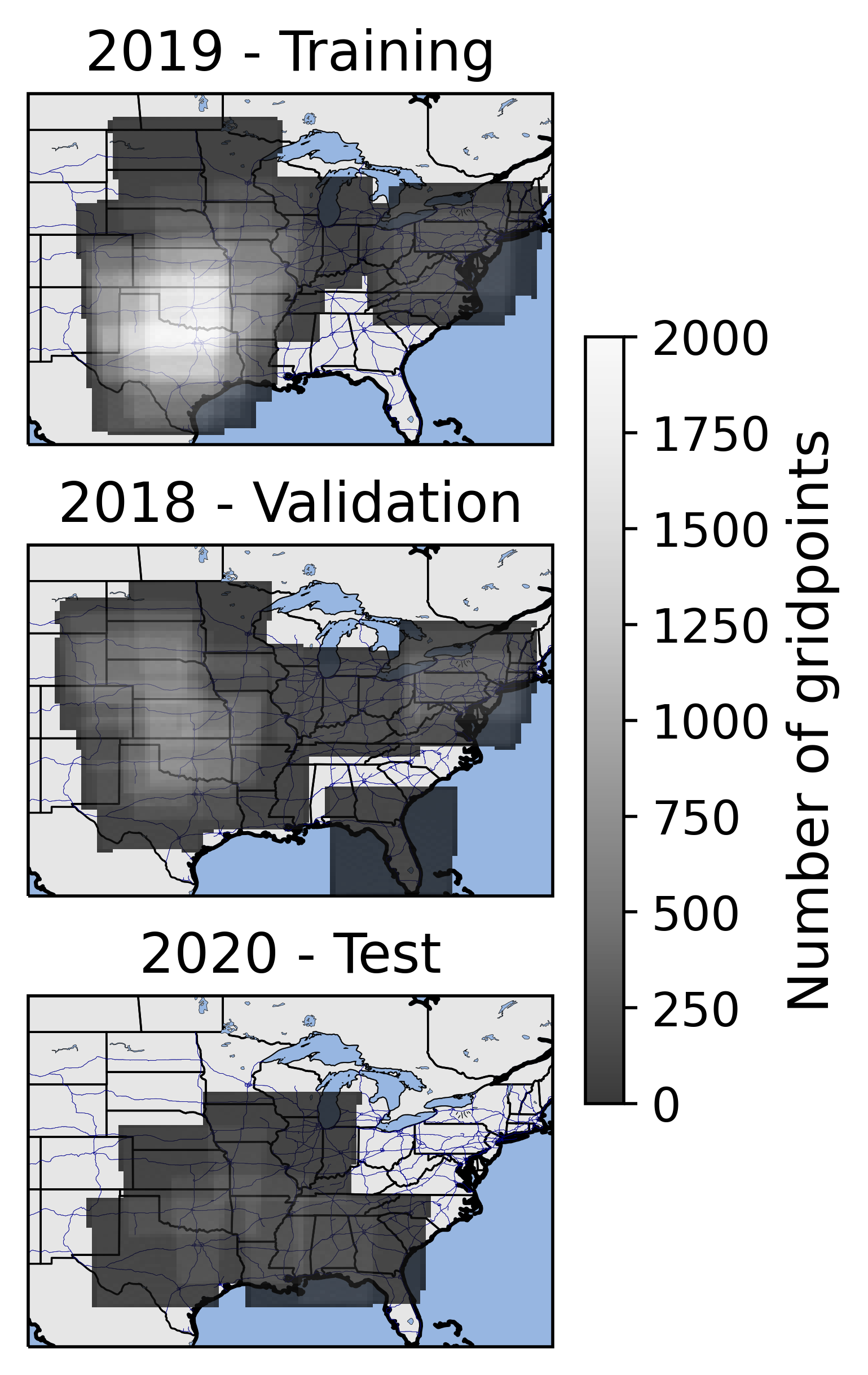}\\
 \caption{Warn on Forecast System domain locations. Top shows all domains where WoFS was run in 2019, middle shows 2018 and bottom shows 2020.} \label{wofs_domains}
\end{figure}

From the random sliced data, a random subset of examples is chosen for the train, validation, and test set. The exact number of examples are $123,209$ ($62\%$), $50,000$ ($25\%$) and $25,000$ ($13\%$) from years 2019, 2018 and 2020 for the train, validation and test set respectively. The year breakdown was selected according to the amount of data that are available: 2019 has the most data, followed by 2018 and 2020 (Fig. \ref{wofs_domains}). Furthermore, saving 2020 for the test dataset enables the most diverse geographic spread of simulations. Note, there is no overlap of data between the train, validation or test set.

The amount of data even after subsetting into the above splits is too large to fit into Random Access Memory (RAM) all at once (e.g., training data is 160 GB, the computers used only have a max of 32 GB). Thus, additional efforts are made to make sure the data are only loaded in \textit{as needed} (i.e., on a batch basis). To accomplish this, the reflectivity data is first normalized using a min-max scaler determined from the training split. Then the entire dataset is converted from float32 precision to float16 precision and saved out to tensorflow datasets. This enables the use of a dataset larger than available RAM on computing resources available. Furthermore, using tensorflow datasets anecdotally resulted in higher stability during training and less crashes of the GPU hardware. Hardware and time of compute is mentioned in the following Section 4.a and Table 1.  

\subsubsection{GridRad}
Since the machine learning model expects 3 km horizontal spacing images, the GridRad data must be resampled from its original 1.5 km spacing. To do the resampling a nearest neighbor and k-dimensional tree (kd-tree) approach is employed. The machine learning model also requires the height coordinate to be defined as above ground level. The default height coordinate definition in GridRad is height above mean sea level. While running the GridRad data with mean sea level data likely works well where ground level elevation is close to sea level (i.e., Florida), there were considerable issues running the model with storms over high terrain (e.g., Wyoming). Thus to convert from mean sea level to above ground level, the ground elevation at every GridRad point is found using the Shuttle Radar Topography Mission (SRTM) project \citep{Farr2007}. From there, the ground elevation is subtracted from the height profile at each pixel. Then a simple linear interpolation was conducted to get the expected heights the machine learning was trained on (i.e., 0.5 km to 17 km above ground level). 

\subsubsection{Dual-Doppler}
In order to make a direct quantitative comparison between the dual-Doppler estimates and the machine learning method all three products were scaled to the same horizontal grid. The process was to first resample the SAMURI dual-Doppler analysis such that it is evaluated on the same exact grid as the CEDRIC dual-Doppler analysis. The resampling is done using a nearest neighbor approach implemented using kd-trees. From the CEDRIC grid, the data were then smoothed to 3-km horizontal grid-spacing, which is the grid-spacing of the machine learning product, using the mean. From there the machine learning method is then resampled from the GridRad grid to the 3 km CEDRIC/SAMURAI grid using a nearest neighbor approach and kd-trees. 

\subsection{Machine Learning Method}

The main machine learning method used in this paper are U-Networks \citep[U-Net; ][]{Ronneberger2015}. U-Nets are well suited for the task of image-to-image translation, which in this paper is the translation of the WoFS simulated reflectivity to the WoFS maximum vertical velocity. Two variants of U-Nets are used in this paper, the standard U-Net \citep{Ronneberger2015} as well as the upgrade to U-Net, the U-Net3+ \citep{Huang2020}. The main addition to the U-Net3+ method is the addition of full-scale feature connections which add skip connections to all convolutions blocks of the U-Net. Code to build the U-Nets can be found in the data availability section. 

Standard loss functions associated with regression tasks include mean squared error or mean absolute error. While these losses can be suitable for some machine learning tasks, they often lead to blurred results when used in image tasks \citep[c.f., Fig. 1 in][]{Ravuri2021}. Furthermore, when using mean squared error or mean absolute error there is no source of uncertainty estimate provided. Thus, a parametric regression loss function \citep[e.g.,][]{Barnes2023} is used here. 

Parametric regression is where the machine learning model outputs parameters of a distribution (e.g., the mean [$\mu$], and standard deviation [$\sigma$] for the normal distribution) as opposed to deterministic predictions. What this enables is probabilistic predictions like the median, 75th percentile, the interquartile range etc. A recent meteorological example of this technique is \citet{Barnes2023}, where parametric regression is used with a multi-layer perceptron network for hurricane intensity prediction.

\begin{figure}[t]
 \centering
 \noindent\includegraphics[width=2.5in]{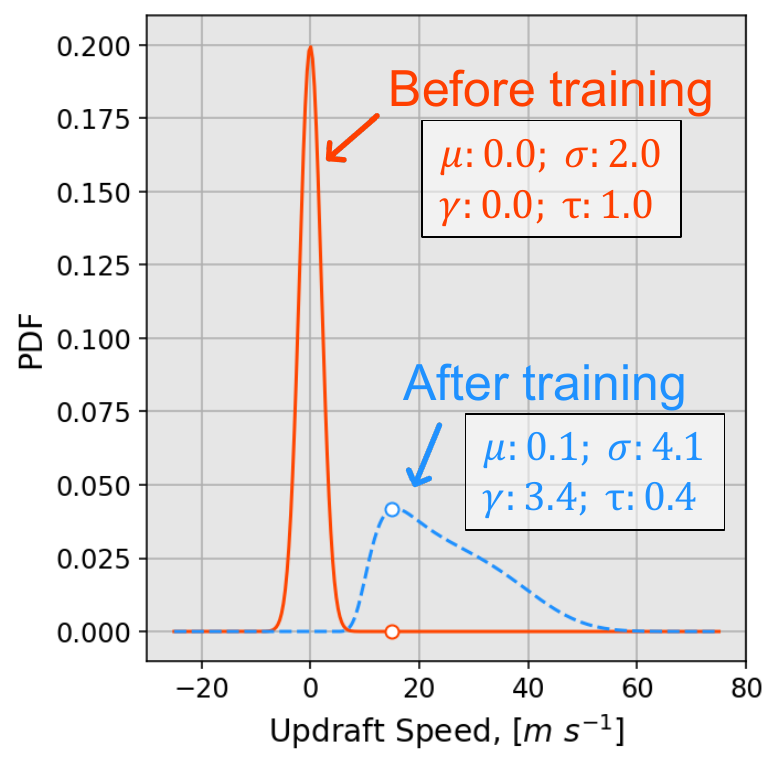}\\
 \caption{Training distributions schematic. The red line depicts a gridpoint at the beginning of training. The circle marker at 18 $\mathrm{m \ s^{-1}}$ is the assumed true updraft value for that pixel. The blue line depicts the distribution after training. Example parameter values are noted in the boxes.} \label{parametric_reg}
\end{figure}

More specifically, \citet{Barnes2023} use a flexible distribution called the sinh-arcsinh-normal distribution, or SHASH. The SHASH distribution can be described by four parameters: location ($\mu$), scale ($\sigma$), skewness ($\gamma$) and tailweight ($\tau$). These four parameters are then the four output channels (i.e., maps) of the U-Net which enable the calculation of parameters like the median vertical velocity. The loss function used for the SHASH method is the negative log likelihood 
 \begin{equation}
    \mathrm{loss} = - \log_{e}(p) \label{e1}
\end{equation}
where p is the value of the SHASH distribution at the target value. We refer readers to \citet{Barnes2023} for more details on the SHASH distribution. 

To exemplify this method, consider Fig. \ref{parametric_reg}. Before training, some grid point sent through the neural network might predict the four red parameters which result in the red curve using the SHASH equation, where the highest likelihood updraft is 0 $\mathrm{m \ s^{-1}}$. The objective of the loss function is to maximize the likelihood of the true value (i.e., the label, which is 18 $\mathrm{m \ s^{-1}}$ for this case). After training, the neural network would ideally produce the four parameters in blue that result in the blue curve where the  maximum likelihood is now near the observed updraft speed of 18 $\mathrm{m \ s^{-1}}$.

For hyperparameter tuning here, some small adjustments to the SHASH method are made. In the original formulation, the $\sigma$ and $\tau$ parameters of the SHASH distribution are positive definite (i.e., cannot be negative). To ensure this property, the output neurons (which are maps here for a U-Net) that represent $\sigma$ and $\tau$ , $\hat{y}_{2}$ and $\hat{y}_{4}$ respectively, are added to the exponent of $e$,

\begin{equation}
    \sigma = e^{\hat{y}_{2}} \label{e2}
\end{equation}
\begin{equation}
    \tau = e^{\hat{y}_{4}}. \label{e3}
\end{equation}
While taking the exponent of the output neurons worked for \citet{Barnes2023}, often in training the initial parameters of the U-Net, initialized using the glorot uniform initialization, would result in $\hat{y}_{2}$ and $\hat{y}_{4}$ sufficiently large that when exponentiated become too large to be represented by the machine precision resulting in not a number (i.e., NaN) and halted training. An empirical solution to this is to reduce the slope of the exponential relation in Equation \ref{e2} and \ref{e3}: 

\begin{equation}
    \sigma = (e^{\hat{y}_{2}})^{\frac{1}{10e}} \label{e4}
\end{equation}
\begin{equation}
    \tau =  (e^{\hat{y}_{4}})^{\frac{1}{10e}}. \label{e5}
\end{equation}
The $10e$ root power is somewhat arbitrary, but is motivated by keeping values roughly in single precision range ($10^{-6}$ to $10^{6}$).

A second adjustment of the SHASH method is an alteration to the loss. For some instances the likelihood of the label (i.e., truth) was very small (i.e., see the red marker and line in Fig \ref{parametric_reg}), thus the negative log likelihood of that value became large in magnitude. The sum of many large loss values (i.e., the sum of the loss across the images in the batch) can lead to NaN loss values. To combat this issue, a small value ($\epsilon$) is added to the loss equation such that the likelihood of p never becomes too small. 

 \begin{equation}
    \mathrm{loss} = - \log_{e}(p + \epsilon) \label{e6}
\end{equation}
where epsilon is $10^{-7}$. The use of $\epsilon$ is commonly done in machine learning to prevent the division by $0$ (i.e., infinity values).

An additional empirical enhancement to the SHASH method comes from the implementation of a weighted loss function. In meteorology, many machine learning tasks of interest are rare (e.g., tornadoes, hail etc.). Even in the context of the number of pixels that have convective updrafts (i.e., greater than 5 $\mathrm{m \ s^{-1}}$) are far out weighted by pixels near zero. This biases the network toward lower updraft magnitudes. To discourage the model from focusing on the weak updrafts a weighted version of the log likelihood is implemented. Specifically, pixels with an WoFS updraft speed greater than a threshold are weighted differently than those pixels with an updraft speed less than that threshold. Both the weights and the thresholds are static hyperparameters. A weight matrix is created and has the same shape as the loss tensor, where the weight value for any pixel is determined by the WoFS value and the total loss is the element wise product between the weight matrix and the loss tensor. Creating a custom loss function with weighting is inspired by \citet{Ebert-Uphoff2021}. 

The previous improvements are to the SHASH method directly. It is also important to consider how the data are scaled before using SHASH. An anecdotal best practice with the SHASH method from this work is to use min-max scaling. The min-max scaling forces the input data into a range of 0-1, which results in more reasonable parameter estimates, especially for the first gradient descent updates, compared to other scaling techniques (e.g., standard scalar) and leads to a more stable hyperparameter search. 

\subsection{Hyperparameters}
With deep learning the number of hyperparameters can be vast and there are no default parameter sets that are guaranteed to achieve satisfactory results on all machine learning tasks. Thus, a random hyperparameter search is conducted. The following parameters are varied: convolutional kernel size, the number of convolutional filters, the depth of the U-Net, the optimizer, batch normalization, batch size, weight of weighted loss function, the threshold for weighting in the weighted loss function and the amount of regularization on the convolutional kernels. For the actual values tested see Tables A1-A3 in the appendix. A total of 100 random model configurations are trained per experiment (see next section). All models are allowed to train for up to 200 epochs, but early stopping is setup such that model training ends if overfitting is detected at which point the best weights (smallest loss) are saved. Models rarely trained past 100 epochs, with a majority training for less than 50 epochs.

\subsection{Model Experiment Setup}
Timeliness is a critically important factor for forecaster tool development \citep{Harrison2022}. Thus, an experiment is setup to test out three different types of machine learning updraft models with varying complexity and thus inference speeds. Experiment one is to train a 2d convolutional U-Net3+ with the composite reflectivity alone (\textit{2dmax}). The goal behind this model search is to emphasize speed such that the updraft product can be provided to the end user in the fastest time possible. A composite reflectivity input should require the least pre-processing time and the 2d convolution U-Net3+ with one feature should produce the fastest inference time. Experiment two is to train a 2d convolutional U-Net3+ on the full 3d reflectivity data (i.e., all 24 levels; \textit{2d24f}). This should be the next fastest model for producing predictions. The last model tested is a 3d convolutional U-Net using the full 3d reflectivity data as inputs (\textit{3d}). The full 3d convolutional U-Net should be the slowest. Each one of the three experiments has 100 hyperparameter configurations. The best model from each hyperparameter search is chosen based on the coefficient of determination ($R^{2}$) between the median of machine learning predicted distribution and the WoFS simulated updraft speed in the validation set.

Beyond the deep learning experiments described above, we also test a simple linear regression model (linreg) to confirm the complicated models (i.e., CNNs) are needed to do the updraft retrieval. It is good practice to compare simple machine learning methods \citep[e.g.,][]{Chase2022} to more complicated methods \citep[e.g.,][]{Chase2023} because simpler models are more understandable by end users and therefore tend to be more trustworthy \citep[][]{Rudin2019}. We train a linear regression on composite reflectivity and only for values greater than 30 dBZ to prevent the linear regression from predicting updrafts for all radar echoes. Because the training and validation datasets are so large, we also train the linear regression off of the test set (i.e., 2020). This will give an unfair advantage to the linear regression, but the resulting performance discussed in the next section will show that the linear regression performance is considerably worse than the deep learning experiments even with the advantage of being trained on the test set.  

\subsection{Evaluation metrics}
A robust evaluation of machine learning is one that uses multiple metrics. While the best models from the hyperparameter search are determined objectively with $R^{2}$, the additional metrics used for evaluation are: the more common statistical definition of $R^{2}$ (which is reported in Table 1), root mean squared error (RMSE), conditional root mean squared error (cRMSE), probability integral transform D statistic (PITD), interquartile range hit rate (IQRr), and intersection over union (IoU). RMSE is defined as 

 \begin{equation}
    \mathrm{RMSE} = \sqrt{\frac{1}{N} \sum_{i=0}^{N} (y_{i} - \hat{y}_i})^{2} \label{e7}
\end{equation}

where $y_i$ is the WoFS maximum vertical velocity for some pixel $i$, $\hat{y}_i$ is the median output from the machine learning method for the same pixel and $N$ is the total number of pixels. The cRMSE is defined as

 \begin{equation}
    \mathrm{cRMSE} = \sqrt{\frac{1}{N} \sum_{i=0}^{N} (y_{i} - \hat{y}_i})^{2} \rvert_{y_i \ge \alpha} \label{e8}
\end{equation}
where the equation is the same, but only evaluated for pixels $i$ that have a $y_i >$ some threshold $\alpha$. IoU is defined as 
 \begin{equation}
    \mathrm{IoU} = \frac{A \cap B}{A \cup B} \rvert_{y_i > \alpha} \label{e9}
\end{equation}
where $A$ is the set of pixels where the WoFS maximum vertical velocity is greater than some threshold $\alpha$ and $B$ is the set of pixels where the machine learning median prediction is greater than the same threshold $\alpha$. An ideal value of IoU is 1 while values greater than 0.5 are generally considered good. These metrics in general give a numerical value for the pixel level accuracy (RMSE), the conditional pixel level accuracy (e.g., for only updraft pixels, how good does the model do; cRMSE) and how well do the machine learning updrafts overlap with the real updrafts (IoU).

In order to evaluate the probabilistic information provided by the SHASH method we employ several metrics. The first metric is the IQRr. This is defined as the rate at which the true value (i.e., WoFS updraft maximum) lies within the interquartile range (25th percentile to 75th percentile). By definition the IQR should contain 50$\%$ of the data, thus an ideal IQRr is 0.5. As a more generalized IQRr, we use additional metrics called the probability integral transform (PIT) histogram and PITD. The PIT value (used in the histogram) is the quantile of the observed value within the machine learning predicted distribution.  In other words, for each quantile range (e.g., 0 – 0.1), how frequent does the observed value fall within this quantile range within the machine learning predicted distributions. Here we binned the PIT values in 0.1 increments (arbitrary choice) and visualized as a normalized histogram. Ideally, the histogram will be flat with all bins having a frequency of the quantile bin width (e.g., 0.1). To quantify how flat the histogram is, the PITD statistic measures the mean deviation from the constant slope, defined as:

 \begin{equation}
    \mathrm{PITD} = \sqrt{\frac{1}{B} \sum_{k=1}^{B}(b_k - \frac{1}{B})^2} \label{e10}
\end{equation}

where $B$ is the number of bins and $b_k$ is the relative frequency of the $k^{th}$ bin. An ideal value of PITD is 0. For more information and examples of the uncertainty metrics please see \citet{Haynes2023} and \citet{Barnes2023}.

As an effort of transparency, we also quantify the time it takes to get a machine learning prediction. These timings are solely done on how long it takes to get an updraft prediction from already available and loaded in radar data. We run 30 batches of size 32 through the model and take the mean time. The number of images in the batch (32) is chosen to ensure the machine learning makes an inference on a domain larger than the United States (recall each image is about 384 km by 384 km). The CPU time was clocked on a 2019 Macbook Pro with an i7 processor and 16 GB of RAM while the GPU time was clocked on Google Colab using their freely available T4 GPUs. 

\section{Results and discussion}

\subsection{Bulk model assessment on WoFS data}

Training the 300 model configurations (100 per experiment) took more than 500 hours of continuous training on a University of Oklahoma owned computer cluster of NVIDIA A100s. The \textit{2dmax} and \textit{2d24f} models were trained on two A100s while the \textit{3d} was trained on four A100s (40GB RAM per card). The best hyperparameter set for each experiment is chosen using the validation set $R^{2}$ value. All metrics described in the previous section are calculated using the test set (2020; Figure 3) and are summarized in Table 1. 

\begin{table*}[ht]
  \centering
    \begin{tabular}{|l|l|l|l|l|l|l|l|l|l|}
   \hline
Models & $R^{2}$ &  RMSE [$m \ s^{-1}$] &  cRMSE [$m \ s^{-1}$] $\rvert_{y_i \ge 5}$&  PIT D & IQRr & IoU $\rvert_{y_i \ge 5}$ & $\#$ param. & CPU [$ms$] & T4 GPU [$ms$]\\ \hline
\textit{linreg}  & 0.42  & 1.70  & 4.80 & * & * & 0.36 & 2 & * & *\\ \hline
\textit{2dmax}  & 0.66  & 0.75  & 4.45 & \textbf{0.04} & 0.62 & 0.45 & 203,972 & 541 & \textbf{58.8}\\ \hline
\textit{2d24f} & 0.73  & \textbf{0.67} & 3.71 & \textbf{0.04} & \textbf{0.52} & 0.51 & 207,284 & \textbf{535} & 147 \\ \hline
\textit{3d} & \textbf{0.75}  & 0.69 &  \textbf{3.67} & 0.07 & \textbf{0.48} & \textbf{0.51} & \textbf{187,140} & 9620 & 702 \\ \hline 
\end{tabular}
\caption{\label{demo-table} Test dataset statistics. Boldface indicates the best of the four experiments. The stars (*) indicate missing or not calculable. $\#$ param. is the number of trainable parameters for each method.}
\end{table*}

\begin{figure*}[t]
 \centering
 \noindent\includegraphics[width=6.4in]{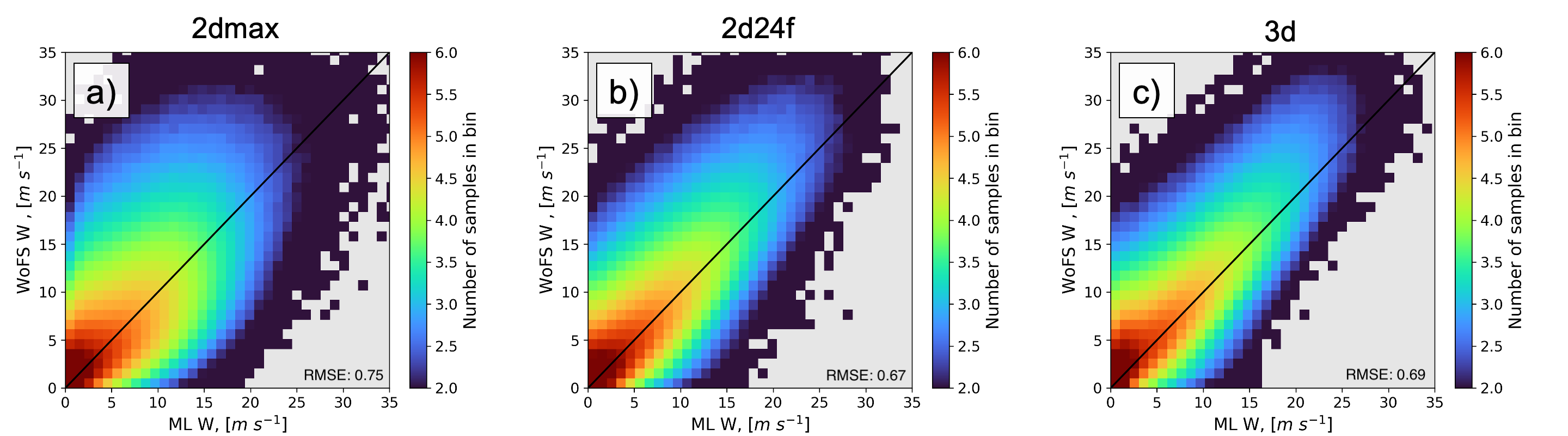}\\
 \caption{One to one comparison of the median machine learning updraft prediction to the WoFS updraft prediction for the test dataset. The statistics for this comparison can be found in Table 1. (a) data for the \textit{2dmax} model (b) data for the \textit{2d24f} model and (c) data for the \textit{3d} model. The colorbar is the log of the counts.} \label{one_to_one}
\end{figure*}

\begin{figure*}[t]
 \centering
 \noindent\includegraphics[width=6.4in]{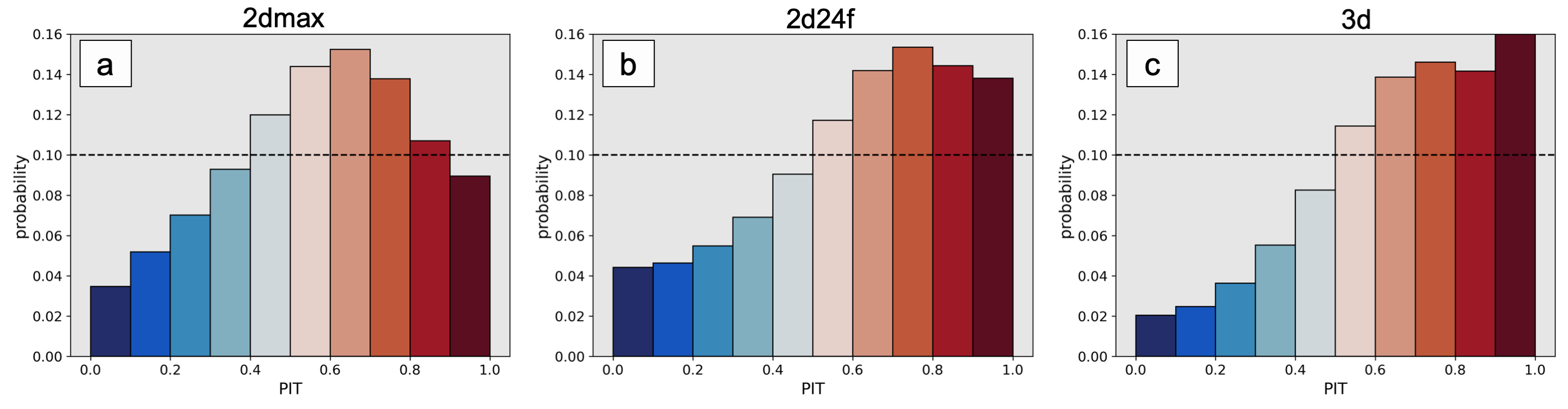}\\
 \caption{Probability integral transform histograms for the (a) \textit{2dmax} (b) \textit{2d24f} and (d) \textit{3d} model evaluated on the test set. The color of the histogram corresponds to the PIT value. The horizontal dashed line is the ideal location of each bar.} \label{PIT}
\end{figure*}

Overall, the experiments (\textit{linreg}, \textit{2dmax}, \textit{2d24f} and \textit{3d}) show expected results, with the deep learning methods performing considerably better than linreg. The \textit{3d} and \textit{2d24f} models have very similar deterministic performance and the \textit{2dmax}, which has less input information, has worse results in all metrics except for the PITD statistic. The deterministic statistics for the deep learning models are visualized in Fig. \ref{one_to_one}, where the pixel wise median from the machine learning predicted distribution is compared directly to the WoFS maximum vertical velocity for the same pixel. If the model prediction was perfect, the output would lie exactly on the diagonal line. The spread about the diagonal is considerably reduced as the model complexity increases (Fig. \ref{one_to_one}a to c). Meanwhile the uncertainty information visualization (i.e., the PIT Histograms; Fig. \ref{PIT}) does not follow the same trend (Fig. \ref{PIT}a to c). The \textit{2dmax} and \textit{2d24f} models have better uncertainty information (PITD of 0.04) than the \textit{3d} model (PITD of 0.07). Beyond the PITD statistic, the \textit{2dmax} model has a peaked histogram at 0.75 PIT while the other two have more left-skewed distributions, with the probability bins of the largest PIT values exceeding 0.16 for the \textit{3d} model. The interpretation of the peaked shape in Fig. \ref{PIT}a, is that the machine learning model is likely under confident in its predictions, predicting too wide of a distribution. Meanwhile the skewed left distributions of Fig. \ref{PIT}b,c is interpreted as the true value landing frequently (more than 10$\%$ of the time) within or beyond the upper tail of the machine learning predicted distribution.

The results of Figs. \ref{one_to_one} and \ref{PIT} suggest that the machine learning predicted distributions are shifted to the left (i.e., closer to 0), which results in the observations being found above the median and at higher quantiles (e.g., 0.75). In general, this indicates an overall low bias in machine learning updraft speed and is supported in the analysis in the following sections. We hypothesize that the low bias is a result of the training data having many pixels where updraft velocities are near 0 $\mathrm{m s^{-1}}$ with non-zero radar reflectivites. We tried to prevent the underestimation with careful curation of the training dataset (Section 3.a.1) and a weighted loss function (Section 3.b) but we were unsuccessful in removing the low bias completely. In the following sections we show that while there is a low bias, the machine learning updraft model has other beneficial qualities, like providing an estimate on updraft width.

The amount of time for inference on the CPU is about an order of magnitude longer between the 2d convolutions (i.e., \textit{2dmax} and \textit{2d24f}) models and the 3d convolutions (i.e., \textit{3d}; Table 1). If a 9620 ms runtime is too large for operational uses, there is the potential to use GPUs in the cloud, which would make all methods run in a similar amount of time. If GPUs are not an available resource, then the user could choose to use the 2d24f method, which has similar skill scores to that of the 3d method but at a faster run time.

It is clear that the \textit{2d24f} and \textit{3d} models are better than the \textit{2dmax} model. Distinguishing if the \textit{2d24f} or \textit{3d} model is better, is more contested (i.e., show similar statistics in Table 1). For the sake of brevity, the remaining figures in the paper will show the \textit{3d} model because of its superior results on the validation set (not shown).

\subsection{A WoFS Example: 30 April 2019}

\begin{figure*}[t]
 \centering
 \noindent\includegraphics[width=6.4in]{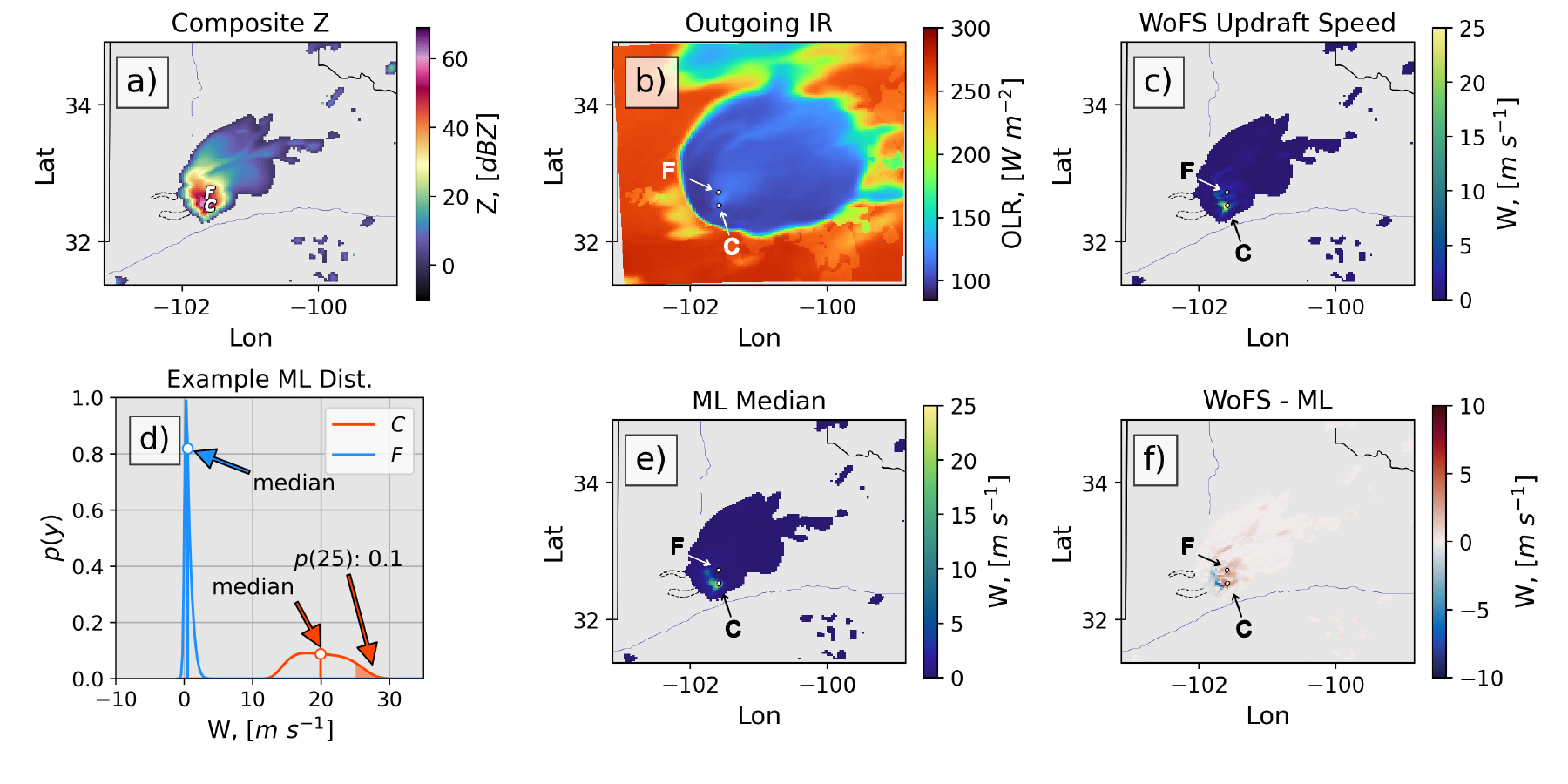}\\
 \caption{A WoFS data example from 30 April 2019 in north Texas. (a) composite reflectivity with a marker of $C$ for the updraft core and the marker $F$ for the forward flank downdraft. The dotted contour is the 1" hail location determined by HAILCAST \citep{Adams-Selin2016}. (b) Simulated outgoing infrared imagery from the same storm. (c) WoFS simulated maximum vertical velocity masked to where the composite reflectivity is greater than 0. (d) Example machine learning distributions from the $C$ and $F$ markers in a. The annotation of p(25) is the integral of the red curve for values greater than 25 $\mathrm{m s^{-1}}$ (e) Median maximum vertical velocity from the machine learning model. (f) Difference between maximum vertical velocity from WoFS and the median prediction from the machine learning.} \label{wofs_case}
\end{figure*}

In order to fully exemplify the machine learning method being used here a case day from the WoFS dataset is used (Fig. \ref{wofs_case}). Note that this case day (all 18 members and initialization times) is not in the training dataset. On 30 April 2019 a shortwave trough was lifting out from New Mexico and headed toward the Oklahoma Texas region. Surface-based convective available potential energy (CAPE) values were forecast to be 3000-4000 $J \ kg^{-1}$ over the Texas panhandle region.  Bulk 0-6 km shear was forecast to be around 50 $kt$ and storm relative helicity (SRH) was forecast to be in the 200-350 $m^{2} \ s^{-1}$ range, supportive of supercells. By the end of the day multiple hail reports were produced by several supercells that traversed northern Texas between Abilene and Amarillo. The tornadoes that occurred on this day were located in eastern Oklahoma, Missouri and Arkansas. 

WoFS member one, initialized at 2000 UTC, has a single supercell traversing the Texas panhandle (Fig. \ref{wofs_case}). Simulated composite reflectivity values exceed 60 dBZ (Fig. \ref{wofs_case}a), outgoing infrared shows an evident spreading anvil (Fig. \ref{wofs_case}b) and WoFS simulated maximum vertical velocities exceed 25 $\mathrm{m \ s^{-1}}$ (Fig. \ref{wofs_case}c). The median machine learning prediction is plotted in Fig. \ref{wofs_case}e. Overall, the machine learning model does well, capturing the primary updraft location to within $\pm$ 5 $\mathrm{m \ s^{-1}}$ (approximately 20$\%$ error) of the WoFS simulated value. 

While the median of the machine learning distribution is useful here (Fig. \ref{wofs_case}e), the full distributions can be interrogated (Fig. \ref{wofs_case}d). Two locations from within the supercell, the core ($C$; Fig 7a) and the forward flank downdraft ($F$; Fig. 7a), exemplify these distributions. The core updraft location shows a broad distribution with non-zero probability density function (PDF) values from about 12 $\mathrm{m \ s^{-1}}$ to 30 $\mathrm{m \ s^{-1}}$ while the forward flank has a steeply peaked distribution with a peak value just above 0 $\mathrm{m \ s^{-1}}$. For this case, the machine learning estimated likelihood of the WoFS simulated updraft maximum (or greater) is 10$\%$. Thus, while the median prediction underestimates the WoFS updraft, the true value is within the predicted distribution. Another encouraging result is that the likelihood of strong vertical velocities in the forward flank, where downdrafts are expected to dominate, is effectively 0. 

\subsection{Transfer Domain Case: 26 May 2017}

\begin{figure*}[t]
 \centering
 \noindent\includegraphics[width=6.33in]{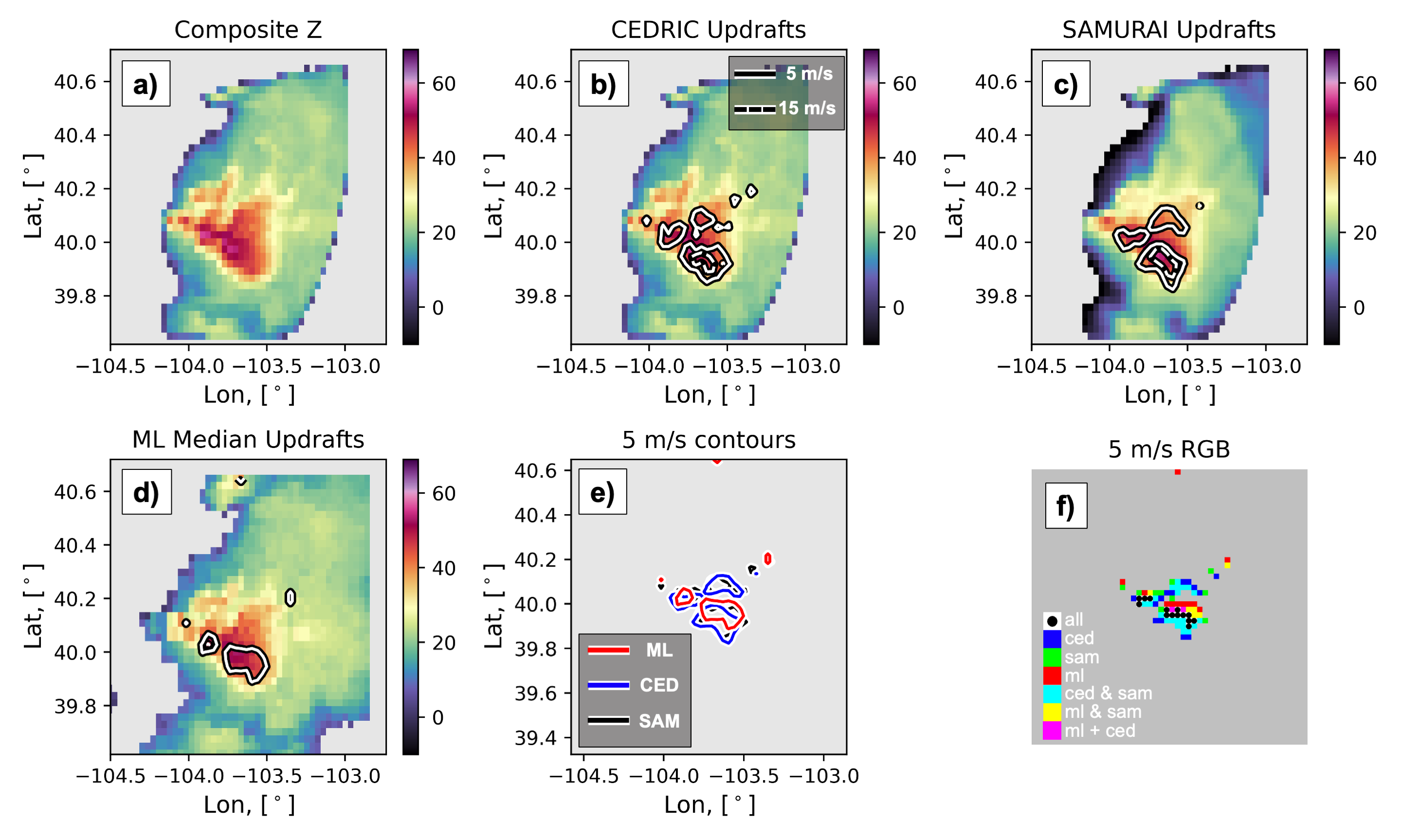}\\
 \caption{A supercell case study from Eastern Colorado on 26 May 2017. (a) observed composite reflectivity from the CEDRIC analysis (b) same as a, but with column maximum vertical velocity contours of 5 and 15  $\mathrm{m \ s^{-1}}$. (c) as in b, but with SAMURAI. (d) as in c but the median machine learning output. (e) all 5 $\mathrm{m \ s^{-1}}$ contours from the CEDRIC (blue) SAMURAI (black) and the machine learning median (red). (f) an RGB image where the color value is determined by if the updraft retrieval has more than 5 $\mathrm{m \ s^{-1}}$ updrafts. The red channel of the RGB image is from the machine learning model, the green channel is from the SAMURAI analysis and the blue channel is from the CEDRIC analysis. Pixels where all methods overlap are in white with a black circle marker in them.} \label{dualdop_case}
\end{figure*}

Thus far the comparisons and evaluations of the machine learning model have been restricted to the WoFS data domain. Note that NWP data is not reality and can be considerably different from observations \citep[e.g.,][]{Varble2014,Fan2017,Marinescu2016}. Furthermore, the motivation of this paper is to train a machine learning model that could be used in real time on observed radar data. Thus as an out-of-sample test case, the 26 May 2017 dual-Doppler case from \citet{Marinescu2020} is analyzed here.

On this day, terrain-initiated convection formed in an environment with surface-based CAPE above 1500 $\mathrm{J \ Kg^{-1}}$ and SRH greater than 200 $\mathrm{m^2 \ s^{-2}}$, supportive of supercells. A lone supercell formed in the dual-Doppler lobe between KFTG and CHILL, allowing for dual-Doppler syntheses of maximum vertical velocity. The closest analysis time (2144 UTC) with the storm closest to the radars for which the dual-Doppler synthesis errors should be minimized is shown in Fig. 8. Both dual-Doppler analyses (i.e., CEDRIC and SAMURAI) show three broad areas of 5 $\mathrm{m \ s^{-1}}$ updrafts around the core of reflectivity, with the strongest updrafts found to the south, with values exceeding 15 $\mathrm{m \ s^{-1}}$ (max of 24 $\mathrm{m \ s^{-1}}$ and 26 $\mathrm{m \ s^{-1}}$ for CEDRIC and SAMURAI respectively; Fig. \ref{dualdop_case}b,c). Meanwhile the machine learning median output captures the two southernmost 5 $\mathrm{m \ s^{-1}}$ updraft regions. The most intense vertical velocity regions determined by the dual-Doppler are generally not captured in the machine learning median output, and the same southern updraft region only shows the 5 $\mathrm{m \ s^{-1}}$ contour (Fig. 8d). The 5 $\mathrm{m \ s^{-1}}$ contours from the three analyses partly overlap (14 pixel overlap; Fig. \ref{dualdop_case}ef). Furthermore, the shapes of the updraft regions, particularly the more elongated shape of the southernmost updraft region is shown in both the dual-Doppler and machine learning data.

\begin{figure*}[t]
 \centering
 \noindent\includegraphics[width=6.4in]{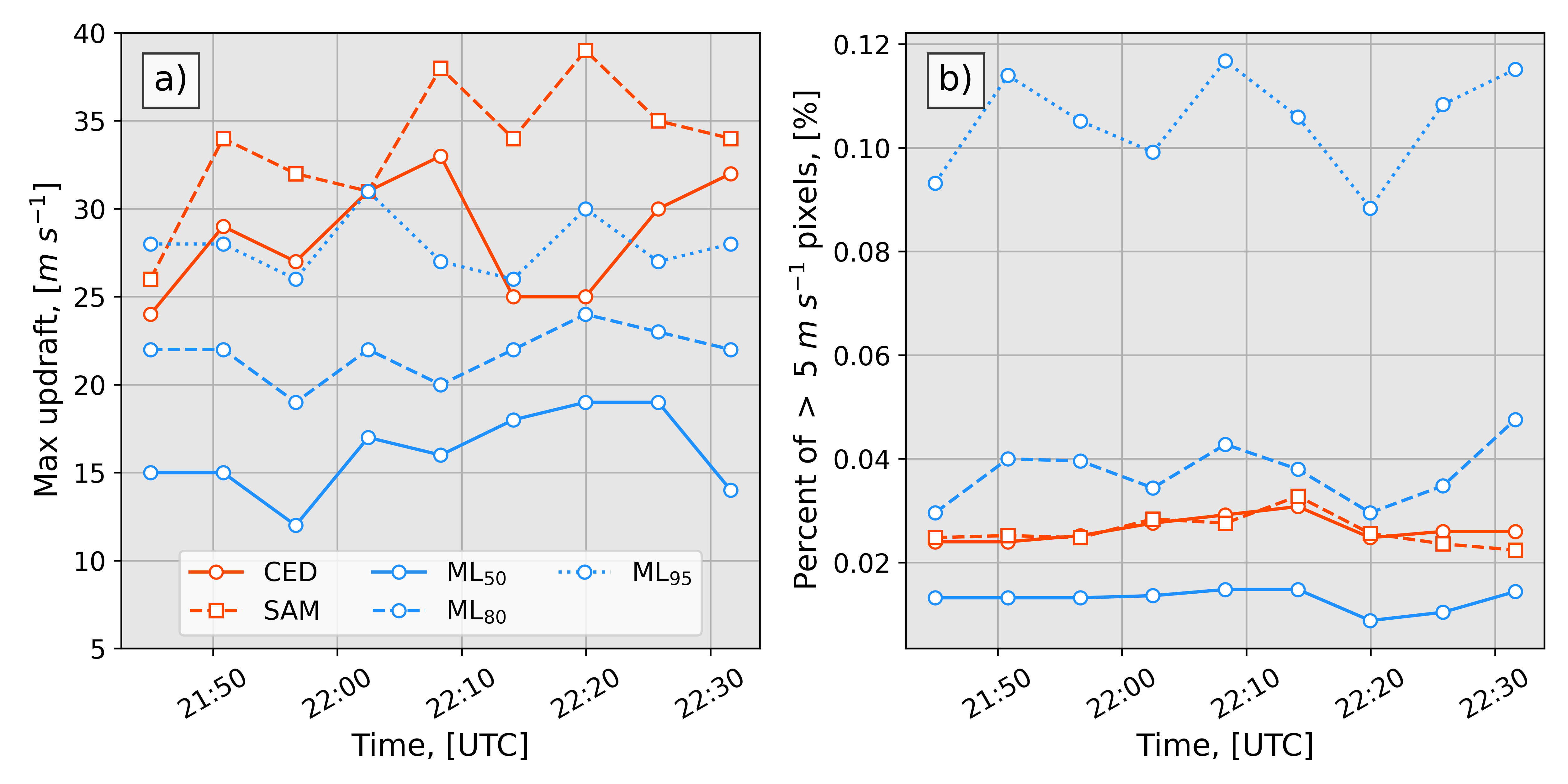}\\
 \caption{Updraft statistics for the 26 May 2017 case. (a) Maximum updraft speed for the main updraft through time. Red lines are the dual-Doppler techniques (dashed SAMURAI, solid CEDRIC), while the blue lines are different quantiles of the machine learning distribution (solid median, dashed 80th percentile, dotted 95th percentile). (b) Percentage of pixels in the entire domain that exceed 5 $\mathrm{m \ s^{-1}}$. Line colors are the same as in a.} \label{max_time}
\end{figure*}

Extracting the max updraft value for the southernmost updraft over time shows that the median prediction from the machine learning is about half of what the dual-Doppler syntheses are suggesting (Fig. 9a). Given the flexibility of the distribution predictions from the machine learning more than the median prediction can be considered (e.g., 80th percentile and the 95th percentile). For this supercell case the 95th percentile resembles the CEDRIC synthesis better (solid and dotted lines Fig. 9).

Part of the motivation of this work is to see if the machine learning could estimate updraft widths properly because the updraft width is important for some severe weather hazards \citep[e.g., tornadoes and hail,][respectively]{Marion2019,Kumjian2021}. We use two evaluations of the updraft widths, a simple quantification of updraft pixels and a more complicated IoU statistic. The simple way to quantify the updraft area is to take the percentage of pixels with vertical velocities greater than 5 $\mathrm{m \ s^{-1}}$ (out of a total of 2500 pixels; Fig. 9b). Note that taking the area of the column maximum vertical velocity has the added ambiguity for tilted updrafts but is still a first order estimation of storm updraft area. Similar to Fig. 9a, the median value of the machine learning distribution is under predicting the total area of 5 $\mathrm{m \ s^{-1}}$. In contrast to Fig. 9a, where larger percentiles capture the dual-Doppler synthesis, the larger percentiles of the machine learning distribution leads to a systematic overestimation of 5 $\mathrm{m \ s^{-1}}$ updraft area. 

The more complicated IoU statistic evaluates if the updraft is in the right location. Thus, the IoU is calculated for each time step and several thresholds of updraft speed (5, 10 and 15 $\mathrm{m \ s^{-1}}$) using only the median of the machine learning distribution. The value of IoU as a function of time is shown in Fig. 10. In general the IoU value between the two dual-Doppler methods is at about 0.5 for all updraft thresholds (Fig. 10a). Meanwhile, the machine learning median predictions shows considerably worse IoU scores compared to CEDRIC (SAMURAI was similar but not shown), where IoU for 5 and 10 $\mathrm{m \ s^{-1}}$ is on average 0.25 and for 15 $\mathrm{m \ s^{-1}}$ IoU can range between 0 and 0.15. The IoU for the machine learning compared to each of the dual-Doppler analyses is about half that of the two dual-Doppler analyses. Since the dual-Doppler techniques have the same input data and follow similar concepts, it is expected that they would have better overlap with each other (Fig. 10a), as compared to the machine learning method (Fig. 10b-c). However, the consistent, albeit smaller, overlap between the machine learning method and dual-Doppler methods, particularly at the weaker vertical velocity thresholds, shows potential skill and application for the machine learning method.

\begin{figure*}[t]
 \centering
 \noindent\includegraphics[width=6.4in]{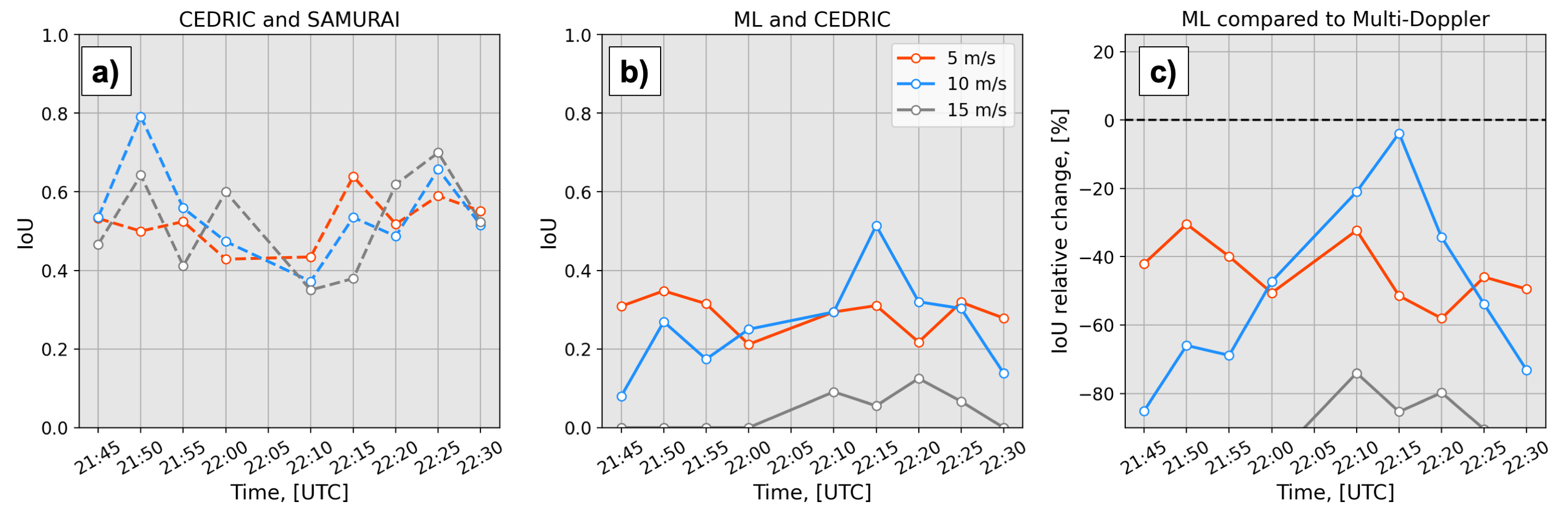}\\
 \caption{Intersection over Union metric through time. Each plot has a 5, 10 and 15 $\mathrm{m \ s^{-1}}$ threshold to binarize the data for the IoU calculation (red, blue and grey respectively). (a) the two dual-Doppler compared to one another (b) the median machine learning model output compared to CEDRIC and (c) the relative difference between panels a and b.} \label{iou_fig}
\end{figure*}

\section{Conclusion}

The severe weather literature notes the importance of a storm's updraft characteristics (i.e., strength and width) in its role in severe weather hazards \citep[e.g.,][]{Trapp2017,Marion2019,Carey2019,French2021,Kumjian2021,Wilson22}. However, only incomplete proxies of storm updraft characteristics exist in real time for forecaster use (see Fig. 1a). Thus, this paper trained a machine learning model, namely a U-Network \citep[U-Net;][]{Ronneberger2015,Huang2020}, to translate 3-dimensional radar reflectivity data into the column maximum vertical velocity (i.e., the maximum updraft) as an effort to create a near real time estimate of storm updraft velocity and area. The U-Net was trained on convective allowing numerical weather prediction data, from the National Severe Storms Laboratory Warn on Forecast System \citep[NSSL WoFS;][]{Stensrud2009,Jones2020}, where the 3-dimensional radar data and column maximum vertical velocity are simulated on the same grid (i.e., same image). Beyond training the machine learning model on WoFS data, an in-depth case study using observations from C\textsuperscript{3}LOUD-Ex was also conducted. The following are the main contributions and conclusions of the paper:

\begin{enumerate}
\item We adapted a parametric regression technique from \citet{Barnes2023} to run with a U-Net, and made several training stability enhancements (Section 3.b). 

\item We showed that a parametric regression U-Net could skillfully reproduce the WoFS updrafts, having coefficients of determination ($R^2$) of $> 0.65$ (best value of 0.75) and root mean squared error (RMSE) on convective updrafts ($> 5 \mathrm{m \ s^{-1}}$) of less than 4.5 $\mathrm{m \ s^{-1}}$ (best value of 3.67 $\mathrm{m \ s^{-1}}$). Furthermore, the model showed skillful updraft area segmentation, characterized by an Intersection over Union (IoU) of greater than 0.45 (best value of 0.51; See Table 1 for more results). 

\item We showed encouraging correspondence between the updraft areas predicted by the machine learning model and those analyzed by two different dual-Doppler techniques in \citet{Marinescu2020}. Machine learning updrafts and the dual-Doppler updrafts characterized by IoU values around 0.25, which is half the IoU of the two dual-Doppler techniques compared to one another. However, the machine learning median updraft magnitude averaged half the dual-Doppler updraft magnitudes.

\end{enumerate}

Overall, the performance of the machine learning model is encouraging given the machine learning model is using only one time step of 3-dimensional radar data. A potential improvement would be to include additional input channels to the machine learning. The goal of this paper was to start simply with only one input field (i.e., reflectivity) where the forward model of numerical weather prediction is generally accepted to be representative of real storm structures. Additional input features to the machine learning method could be previous time steps of reflectivity since there is evidence that the temporal evolution of storm reflectivity structures are related to vertical velocity \citep[e.g.,][]{Haddad2022,Prasanth2023}. If forward simulations can be realistically done of dual-pol parameters from the bulk microphysics of the numerical weather prediction then differential reflectivity (Zdr) and specific differential phase (Kdp) could also be used as inputs given their vertical structures in storms \citep[e.g.,][]{Homeyer2020,Homeyer2023}. Alternatively, simulated Doppler radar moments could be included (e.g., radial divergence, azimuthal shear etc). Another improvement to the technique here would be to re-train the machine learning using higher resolution numerical weather prediction. The use of the three kilometer horizontal grid-spacing is likely a major limitation of what spatial structures of reflectivity the machine learning is leveraging and what vertical motions are resolved \citep[e.g.,][]{Potvin2015,Schwartz2017,Marinescu2020,Squitieri2020}. Thus, models using one kilometer grid-spacing, which is closer to observed gridded radar resolution and is also being tested for WoFS \citep{Kerr2023}, could assist in the representation of reflectivity distributions as well as further resolving more convective motions. Note that all these additions will add to the computational time required to do machine learning inference which shouldn't be forgotten since timeliness is a critical aspect of a weather forecasting tool \citep{Harrison2022}. 

Beyond direct improvements to the machine learning model, more evaluations should be conducted. One such evaluation would include compiling many multi-Doppler cases that span diverse meteorological conditions (i.e., bow echoes, hurricanes etc.) to contextualize the biases in the existing machine learning model. Another evaluation that could be done is to compare the machine learning output directly to storm reports of severe weather hazards. There might exist some thresholds or key patterns that come before a severe weather hazard that could be used to enhance forecasts. Finally, comparing all the proxy methods and the new machine learning method would highlight the strengths of each method, providing the best practices for each method to overall enhance a forecasters ability to diagnose severe storm potential. 

Once the machine learning model is ready for operational use, the transition to operations should be carefully considered. Operational radar data have artifacts like ground clutter and non-meteorological targets that when used as input to the non-linear machine learning model could lead to spurious updrafts or non-physical updraft values. Furthermore, observed radar data often miss the lowest levels of storms (Maddox et al. 2002) which could be problematic for a machine learning model that was trained with WoFS where low-level data were available all the time. Despite these caveats, it is encouraging to see the machine learning models perform well on the single test of observational data in this paper, but further tests should be conducted before operational use.

%

\clearpage
\acknowledgments
This material is based upon work supported by the National Science Foundation under Grant No. ICER-2019758, supporting authors RJC, and AM. We acknowledge the support staff at the OU Supercomputing Center for Education and Research (OSCER) that helped with setting up and maintaining computing facility that enabled this research. Author PJM was provided support by INCUS, a NASA Earth Venture Mission, funded by NASA’s Science Mission Directorate and managed through the Earth System Science Pathfinder Program Office under contract number 80LARC22DA011. The C\textsuperscript{3}LOUD-Ex field campaign and its associated data used in this study were supported by the Monfort Excellence Fund provided to Susan C. van den Heever as a Monfort Professor at Colorado State University, as well as funding from the National Science Foundation Grant AGS-1409686. CKP's contribution to this work comprised regular duties at federally funded NOAA/NSSL.

We thank William McGovern-Fagg for his help editing the NSSL graphic to make it easier to read. We thank Kayla Hoffman for being the first to work on the project as part of the 2022 research experience for undergraduates at the University of Oklahoma. We thank Daniel Stechman, Stephen Nesbitt and Matthew Wilson for their valuable input on this project. We also thank the three anonymous reviewers for their kind words and their thorough evaluation of our manuscript. 


%
%
\datastatement

The data for the test set used here is available for download \url{https://doi.org/10.5281/zenodo.10001880}, the dual-Doppler data can be obtained from the authors of \citet{Marinescu2020}. The trained models and scripts can be found on the github repo associated with this manuscript \url{https://github.com/ai2es/hradar2updraft}. Unfortunately, the training and validation data are too large to host online (more than 200 GB), thus they are available upon request to the corresponding author. The raw WoFS data are considered experimental and not available for use from the corresponding author. Please reach out to NSSL for more information on the WoFS data. 


%
\appendix

\appendixtitle{Hyperparameter Tuning Specifics}
All the models shown in the paper are the result of a fairly extensive hyperparameter search. Each of the following figures contains the different hyperparameters that were varied. Note that a total of 100 models were trained for each model type, so it is very possible that not all possible hyperparameter solution sets were run.

\begin{figure}
\centerline{\includegraphics[width=3in]{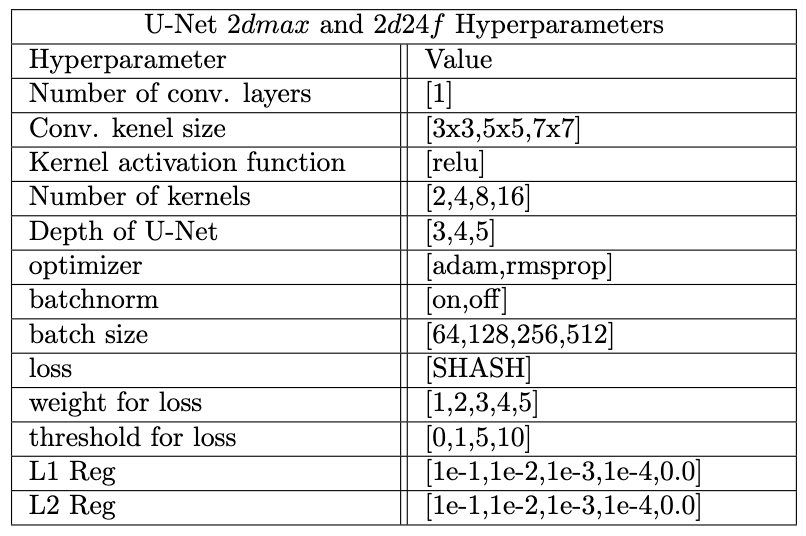}}
\caption{Figure showing the hyperparameters for the $2dmax$ and $2d24f$ experiments}
\end{figure}

\begin{figure}
\centerline{\includegraphics[width=3in]{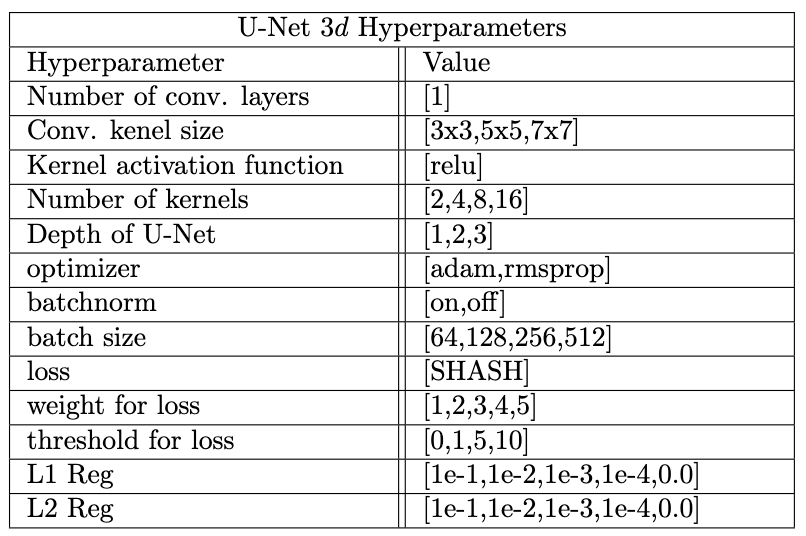}}
\caption{Figure showing the hyperparameters for the $3d$ experiment}
\end{figure}
\clearpage

\bibliographystyle{ametsocV6}
\bibliography{references}

\end{document}